\documentclass{article}
\usepackage{spconf,amsmath,graphicx}
\usepackage{array}
\usepackage{bbding}
\usepackage{amssymb}
\usepackage{mathrsfs}        
\usepackage{graphicx}        
\usepackage{amssymb}         
\usepackage{amsmath}         
\usepackage{enumerate}       
\usepackage{booktabs}
\usepackage[colorlinks,
            linkcolor=red,
            anchorcolor=blue,
            citecolor=blue
            ]{hyperref}

\usepackage[font=small,labelsep=period]{caption}
\usepackage{amsthm}          
\usepackage{algorithm}
\usepackage{algorithmic}
\usepackage{subfigure}
\usepackage{mathrsfs}
\usepackage{graphicx}
\usepackage{amssymb} 
\usepackage{amsmath} 
\usepackage{amsthm}
\usepackage{enumerate}
\usepackage[numbers,sort&compress]{natbib}

\usepackage{amssymb}

\title{MFEViT: A Robust Lightweight Transformer-based Network for Multimodal 2D+3D Facial Expression Recognition}
\name{Hanting Li, Mingzhe Sui, Zhaoqing Zhu, Feng Zhao\thanks{The corresponding author is Feng Zhao}
}
\address{University of Science and Technology of China, Hefei 230027, China\\
    \{ab828658, sa20, zhaoqingzhu \}@mail.ustc.edu.cn, \{fzhao956\}@ustc.edu.cn\\
    }
\begin{document}
%
\maketitle
\begin{abstract}
Vision transformer (ViT) has been widely applied in many areas due to its self-attention mechanism that help obtain the global receptive field since the first layer. It even achieves surprising performance exceeding CNN in some vision tasks. However, there exists an issue when leveraging vision transformer into 2D+3D facial expression recognition (FER), i.e., ViT training needs mass data. Nonetheless, the number of samples in public 2D+3D FER datasets is far from sufficient for evaluation. How to utilize the ViT pre-trained on RGB images to handle 2D+3D data becomes a challenge. To solve this problem, we propose a robust lightweight pure transformer-based network for multimodal 2D+3D FER, namely MFEViT. For narrowing the gap between RGB and multimodal data, we devise an alternative fusion strategy, which replaces each of the three channels of an RGB image with the depth-map channel and fuses them before feeding them into the transformer encoder. Moreover, the designed sample filtering module adds several subclasses for each expression and move the noisy samples to their corresponding subclasses, thus eliminating their disturbance on the network during the training stage. Extensive experiments demonstrate that our MFEViT outperforms state-of-the-art approaches with an accuracy of 90.83\% on BU-3DFE and 90.28\% on Bosphorus. Specifically, the proposed MFEViT is a lightweight model, requiring much fewer parameters than multi-branch CNNs. To the best of our knowledge, this is the first work to introduce vision transformer into multimodal 2D+3D FER. The source code of our MFEViT will be publicly available online.
\end{abstract}

\section{Introduction}
Facial expressions are an essential carrier for spreading human emotional information and coordinating interpersonal relationships. Automatic facial expression recognition (FER) generally aims to recognize six prototypical expressions: anger (AN), disgust (DI), fear (FE), happiness (HA), sadness (SA), and surprise (SU). It has been widely leveraged in many human-computer interaction areas \cite{li2020deep, Guo2020Real, Lian2020Expression}. However, illumination and poses are two unstable factors in 2D FER in that different radiation angles, illumination intensities, and poses may all have a great impact on the image quality and recognition performance \cite{li2020deep}. Therefore, some researchers have attempted to merge 3D depth information that is more robust to the illumination and pose variations with 2D features. Previous deep learning-based approaches \cite{Li2017Multimodal, Zhu2019Discriminative, Lin2020Orthogonalization, Sui2021FFNet} for multimodal 2D+3D FER utilize multi-branch convolutional neural networks (CNNs) to separately extract features for each modality, thus requiring a large number of parameters and also a high cost of memory.

Recently, vision transformer (ViT) has made remarkable progress in image classification tasks \cite{dosovitskiy2020ViT}. Unlike CNN, its strong and unique self-attention mechanism helps obtain the global receptive fields since the first layer, which also makes ViT strongly depend on large amounts of data for pre-training. However, there are no public large-scale 2D+3D FER datasets comparable in scale to the ImageNet \cite{krizhevsky2012imagenet}. How to better transfer the ViT pre-trained on RGB modality to the 2D+3D FER tasks is still a vital challenge. On the other hand, the big intra-class distance for each expression from various subjects is inherent in the FER tasks and it can bring disturbance to the model training. As illustrated in Figure 1, it is quite hard for the network to fit the noisy samples in the right box well. More severely, they may be misclassified into the wrong category. Consequently, it is necessary to consider eliminating the negative effects from the noisy samples in every class during the training process.

\begin{figure}[t]
	\centering
	\centerline{\includegraphics[width=9cm]{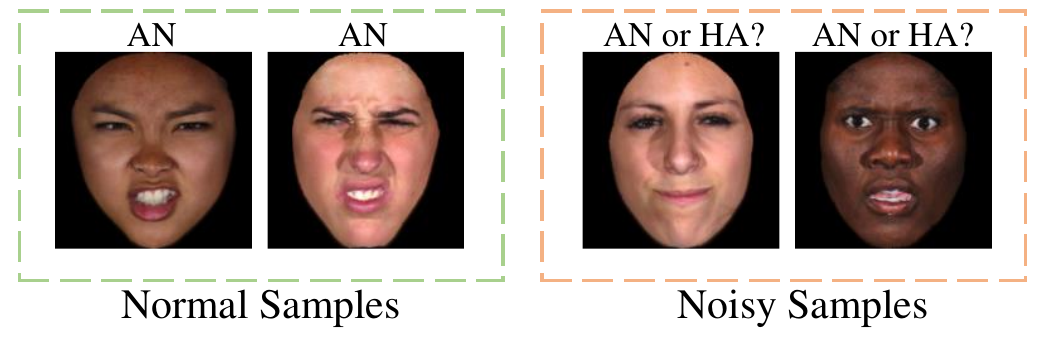}}	
	\caption{Normal samples and noisy samples. All of them belong to the category of anger in original dataset.}
	\label{figure1}
\end{figure}

To address the aforementioned two issues, we propose a robust lightweight pure transformer-based network, named multimodal facial expression vision transformer (MFEViT) for 2D+3D FER. It consists of two primary parts: alternative fusion (AF) strategy and sample filtering (SF) module. In preprocessing, given each original 3D scan, we first convert it into a pair of strictly aligned RGB image and depth map through interpolation and projection, and then apply the surface processing to improve their quality. For extracting multimodal features and classification, the AF strategy first replaces each of the three channels of an RGB image with the depth-map channel and then fuses them before the transformer encoder, which can narrow the gap between RGB and multimodal data to better leverage the ViT pre-trained on ImageNet. Since we perform 2D+3D fusion at the data level, unlike previous multi-branch CNNs, only a single transformer encoder is required for extracting multimodal features, thus reducing the amount of parameters immensely. For reducing the interference of noisy samples on the training of our MFEViT, the designed SF module adds several subclasses for each expression and move the noisy samples to their corresponding subclasses. At the testing stage, we merge all the subclasses into their corresponding main classes and compare them with the ground truths to calculate the overall accuracy.

The main contributions of this work are listed as follows.
\begin{itemize}
	\item We present a robust lightweight pure transformer-based model called MFEViT for multimodal 2D+3D FER. Our MFEViT requires fewer parameters in comparison with existing CNN-based methods and is more robust for dealing with noisy samples.
	\item We propose an alternative fusion strategy to narrow the gap between RGB and multimodal data. Besides, we design a sample filtering module to eliminate the impact of noisy samples on the model training due to the big intra-class distance.
	\item Our MFEViT achieves an accuracy of 90.83\% on BU-3DFE and 90.28\% on Bosphorus with fewer parameters, outperforming other state-of-the-art (SOTA) approaches. To the best of our knowledge, this is the first work to introduce the vision transformer into multimodal 2D+3D FER.
\end{itemize}
\section{Related Work}
\subsection{2D+3D FER}
3D models can capture more tiny and subtle deformations appearing on the face due to the changes in various facial expressions. Merging 2D and 3D features together will present a better performance on FER. According to the feature extraction methods, 2D+3D FER can be roughly divided into two categories: hand-crafted features and deep network-based features. The former leverages traditional feature operator on 3D attribute maps obtained from original 3D scans, such as depth-SIFT features \cite{Berretti2010A}, normal-LBP \cite{Li20123D}, and curvature-HOG \cite{Lemaire2013Fully}. Recently, the techniques based on deep CNN features have fully surpassed the performance of hand-crafted features. Li \textit{et al.} first introduced CNN into 2D+3D FER \cite{Li2017Multimodal}. They used a multi-branch deep CNN network to extract features from the six inputs, one of which is the 2D RGB image and the remaining ones are 3D attribute maps. The feature-level strategy splices the output of each branch to make the final classification. Later, Zhu \textit{et al.} proposed the DA-CNN network based on attention mechanism to enlarge the receptive field and encode richer local clues in each branch \cite{Zhu2019Discriminative}. Similarly, Sui \textit{et al.} utilized facial landmarks as extra prior knowledge to generate masks, which are incorporated to the network and help highlight local features \cite{Sui2021FFNet}. Lin \textit{et al.} thought there exists redundancy between multimodal features and designed the orthogonal loss to reduce the correlation of them \cite{Lin2020Orthogonalization}.

However, the deep networks of almost all CNN-based approaches are multi-branch because the fusion process is performed at the feature level, inevitably demanding mass parameters and high memory cost.
\subsection{Vision Transformer}
Transformers were first proposed by Vaswani \textit{et al.} \cite{vaswani2017transformer} for machine translation, and have become one of the most popular backbones in many natural language processing (NLP) tasks. ViT \cite{dosovitskiy2020ViT} is the first work to apply a transformer directly to a sequence of image patches, which obtains SOTA performance on image classification. The authors concluded that transformers ``do not generalize well when trained on insufficient amounts of data'', which means that ViT relies heavily on pre-training of large-scale datasets, i.e., ImageNet-21K \cite{deng2009imagenet}. To alleviate the problem, data-efficient image transformer (DeiT) was developed to reduce the dependence of ViT on large-scale datasets by utilizing data augmentation and a distillation token \cite{touvron2020DeiT}. Recently, TriTransNet built on ViT was designed to model both RGB and depth features extracted from convolutional neural networks for salient object detection \cite{liu2021TritransNet}, which demonstrates the effectiveness of ViT on 2D+3D multimodal tasks. Inspired by these transformer-based methods, we propose the first multimodal FER framework MFEViT built entirely on transformer, which is a single-branch model since the multimodal information is fused at the data level.
\begin{figure*}[htb]
  \centering
  \centerline{\includegraphics[width=18cm]{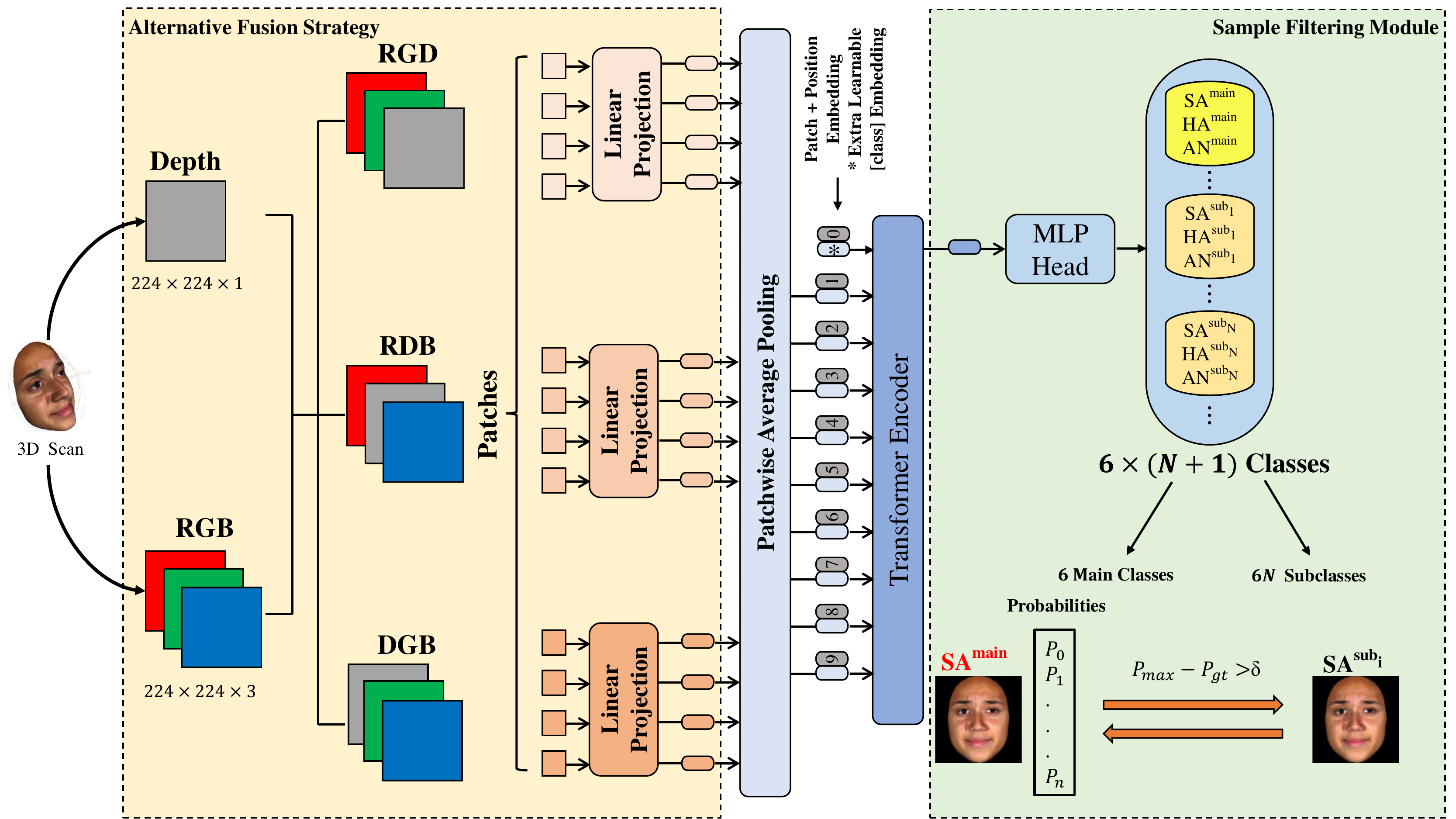}}
  \medskip

\caption{The pipeline of the MFEViT. It contains two crucial parts: alternative fusion strategy to fuse multimodal data and sample filtering module to deal with the noisy samples.  }
\end{figure*}

\section{Method}
To learn robust facial expression features from both depth maps and RGB images, we devise a robust lightweight pure transformer-based framework named multimodal facial expression vision transformer. Unlike previous methods \cite{Li2017Multimodal, Zhu2019Discriminative}, we do not use multi-branch networks to deal with information from different modalities. To fuse 2D and 3D information, we design an alternative fusion strategy, which can narrow the gap between RGB images and multimodal data (2D+3D), and largely reduces the amount of network parameters. Moreover, we introduce a novel sample filtering module to weaken the interference of noisy samples (see Figure 1). To fully demonstrate the advantages of ViT for visual modeling tasks, we retain most of the internal structure of ViT \cite{dosovitskiy2020ViT}. Since training ViT often requires large-scale datasets, we directly use the DeiT-S pre-trained on ImageNet \cite{krizhevsky2012imagenet} in \cite{touvron2020DeiT} as the backbone of our MFEViT. To give the details of our approach, we first briefly introduce the preprocessing for original 3D scans, and then describe the overall architecture of our MFEViT. After that, we present the two key parts of MFEViT in detail.
\subsection{Preprocessing}
Before extracting multimodal features, we first need to get strictly aligned 2D RGB images with three channels and 3D depth maps with one channel. Here, we use the gridfit interpolation \cite{S2018Learning} and projection for each 3D scan to obtain aligned RGB images and depth maps. Then, we apply the surface processing containing three steps \cite{Sui2021FFNet}, namely outlier removal, hole filling, and noisy removal to improve the data quality.
\subsection{Network Architecture}
Our MFEViT is built on a pure transformer and composed of two crucial parts: i) alternative fusion strategy and ii) sample filtering module, as depicted in Figure 2.

The alternative fusion strategy is utilized to fuse the RGB images and depth maps. For each pair of three-channel RGB image I$_{RGB}$ with the size of $224$$\times$$224$$\times$$3$ and single-channel depth map I$_{D}$ with the size of $224$$\times$$224$$\times$$1$, we use I$_{D}$ to replace each of the three channels of I$_{RGB}$, respectively. So, we get three different three-channel representations with the size of $224$$\times$$224$$\times$$3$ as the inputs: I$_{RGD}$, I$_{RDB}$, and I$_{DGB}$, as shown in Figure 2. Then, we reshape them into three sequences of flattened 2D patches with the size of $M\times$$(16^{2}\times 3)$: I$_{pRGD}$, I$_{pRDB}$, and I$_{pDGB}$. Here, $M$$=$$224$$\times$$224/16^{2}$$=$$196$ is the number of patches in each sequence, which also serves as the effective input sequence length for our MFEViT. Following that, we apply linear projection to I$_{pRGD}$, I$_{pRDB}$, and I$_{pDGB}$ to get three sequences of patch embeddings with the size of $M$$\times$$D$: E$_{RGD}$, E$_{RDB}$, and E$_{DGB}$, where $D$ is the channel number of each patch embedding. Subsequently, we attain the E$_{fusion}$ by doing a patchwise average pooling operation on E$_{RGD}$, E$_{RDB}$, and E$_{DGB}$. After concatenating a learnable classification token E$_{class}$ with $D$ dimensions to E$_{fusion}$ and adding position embeddings E$_{pos}$ with the size of $(M$$+$$1)$$\times$$D$, we finally obtain the input of the ViT encoder for classification. To further improve MFEViT, we design a sample filtering module to reduce the harm of noisy samples to the generalization ability of networks. At the training stage, we map E$_{class}$ to a $6$$\times$$(N$$+$$1)$-dimensions vector rather than six dimensions which correspond to the six basic expressions after a 12-layer transformer encoder. Note that the first six units of the output represent the main class of the six basic expressions, and the remaining $6N$ elements stand for the $N$ subclasses of each basic expression that will be employed to accommodate the noisy samples in FER datasets.

\subsection{Alternative Fusion Strategy}
ViT often relies on pre-training on large-scale datasets due to its powerful modeling ability. However, there are no public multimodal FER datasets comparable to AffectNet \cite{mollahosseini2017affectnet}, which currently contains about 450,000 images. Therefore, we propose an alternative fusion strategy to narrow the gap between RGB image and multimodal data. In doing so, our MFEViT can better transfer the model pretrained on large-scale RGB image datasets like ImageNet \cite{krizhevsky2012imagenet} to multimodal facial expression datasets. \\ \indent The RGB images contain some meaningful information \cite{saxena20072Dto3D}, while the original depth maps have almost no 2D texture information. So we utilize the depth maps to supplement the depth information involved in the three channels of the RGB images, as shown in Figure 2. \\ \indent Specifically, we use a single-channel depth map to replace each of the three channels of a RGB image respectively, and feed the three representations of fused data I$_{RGD}$, I$_{RDB}$, and I$_{DGB}$ into three independent encoders to obtain three sequences of patch embeddings. After that, we calculate the average of them to generate the input for ViT. The previous methods try to map a 3D scan into several three-channel pseudo-color images matching the RGB image \cite{zhu2019DA-CNN} in order to directly utilize common backbone networks such as VGG16 \cite{simonyan2014VGG} and ResNet \cite{he2016ResNet} for processing 3D information \cite{Li2017Multimodal,Sui2021FFNet}. Therefore, these approaches usually require multi-branch networks with independent parameters to handle different-modal data and fuse them at the feature level. Undoubtedly, this will greatly increase the network parameters and training cost. By contrast, our MFEViT requires only a single branch because the fusion is done at the data level, which effectively reduces the network parameters and the difficulty of deployment. Compared with CNN, transformer-based models depend more on large amounts of data for pre-training. In other words, it is more difficult to transfer the transformer-based models pre-trained on the RGB datasets to multimodal data. Therefore, our alternative fusion module can also effectively alleviate the problem of the large gap between the multimodal data and the RGB images by retaining two channels of the RGB image and adding one depth channel at a time.

\subsection{Noisy Samples Filtering}

\begin{figure}[t]
  \centering
  \centerline{\includegraphics[width=8cm]{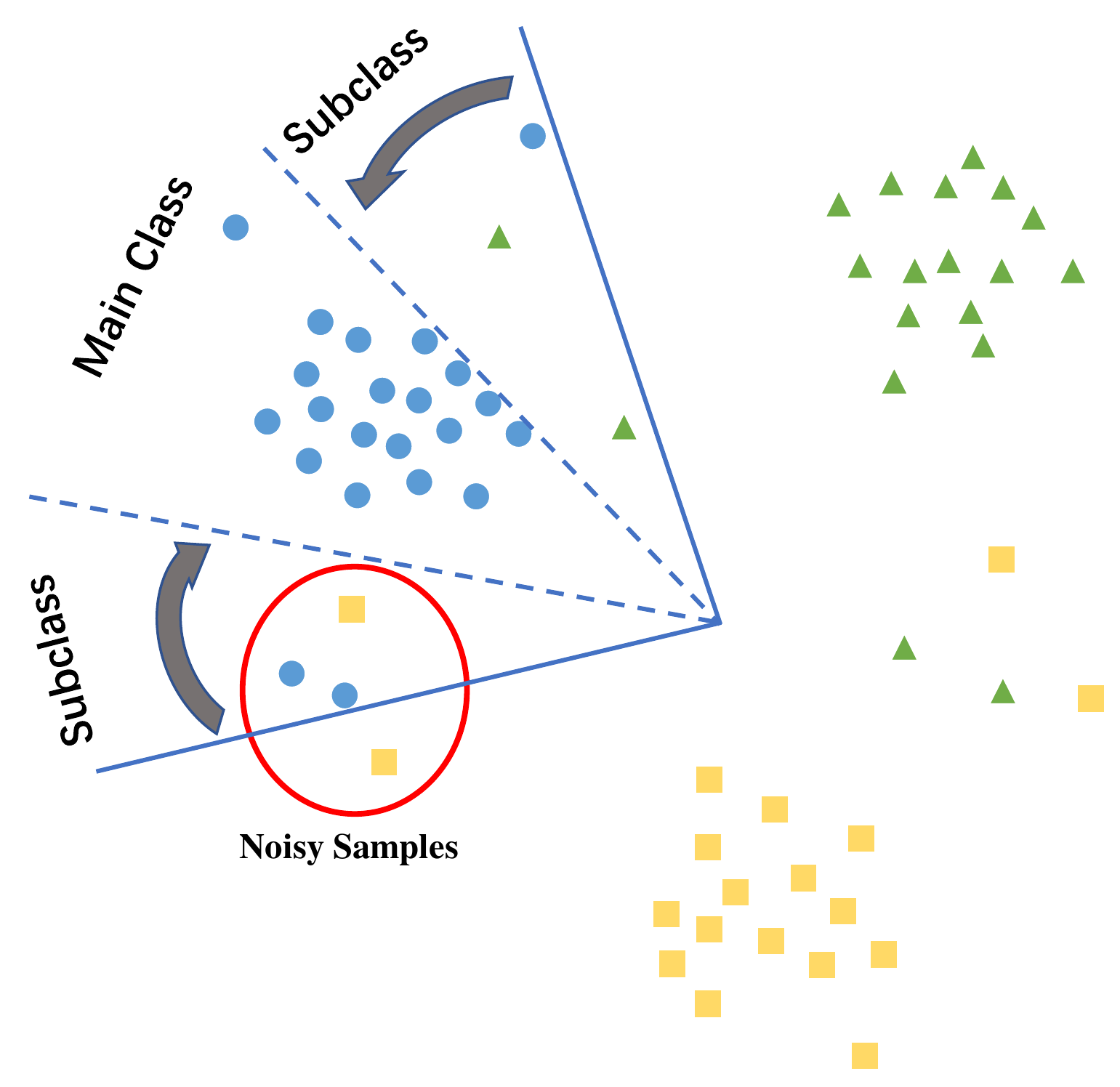}}
  \medskip

\caption{Sample filtering (SF) module. }
\end{figure}
Although the annotations of laboratory FER datasets are more accurate than those of FER datasets in the wild, there still exists big intra-class distance for each expression. As described in Figure 3, if all the blue circular samples are grouped into one category, the margin between the blue circular samples and those of other classes will be reduced, thereby lowering the generalization ability of the network. Hence, we devise the sample filtering module to minimize the influence of noisy samples on the network generalization without affecting the training of normal samples.

The relabeling module in self-cure network (SCN) \cite{wang2020SCN} and mask vision transformer (MVT) \cite{li2021mvt} attempt to correct the wrong labels in 2D FER datasets. Different from them, we assume that the labels of laboratory datasets are accurate, so we will not change the original class labels of the samples (e.g., changing the label of the anger to happiness). As shown in Figure 2, the SF module will only change the label of a noisy sample to the subclass of its originally annotated label, and the normal samples will be kept in the original main class. Specifically, we map the class token used to decide the final expression to a vector with $6$$\times$$(N$$+$$1)$ dimensions. The first six dimensions correspond to the six main classes of the six basic expressions, and the remaining $6N$ dimensions correspond to the N subclasses (1-th to N-th) of the six basic expressions, respectively. For each sample, we compare the maximum predicted probability $P_{max}$ with the probability of the current label $P_{gt}$. Similar to the trigger condition in SCN \cite{wang2020SCN}, the sample will be relabeled if and only if the difference between its $P_{max}$ and $P_{gt}$ is higher than the fixed threshold $\delta$. Formally, the module can be defined as,
\begin{equation}\label{1}\scriptsize
 l_{new}=\left\{\begin{array}{rcl} l_{max1} \quad if P_{max}-P_{gt}>\delta \quad and \quad l_{max1} \neq l_{cur}, \\
 l_{max2} \quad if P_{max}-P_{gt}>\delta \quad and \quad l_{max1}=l_{cur}, \\
 l_{cur} \quad \quad \quad \quad  \quad\quad \quad \quad \quad \quad \quad \quad \quad \quad \quad \quad otherwise,\end{array}\right.
\end{equation}
\noindent where $l_{new}$ denotes the new label, $\delta$ is the threshold that triggers the relabeling operation, $P_{max}$ stands for the maximum predicted probability of all classes, $P_{gt}$ represents the predicted probability of the current label, $l_{max1}$ and $l_{max2}$ are the indices of the maximum prediction and the second maximum prediction of the $N$$+$$1$ classes of the originally annotated expression, respectively, and $l_{cur}$ means the current label. In particular, when $l_{max1}$ and $l_{cur}$ are the same, we will change the label to $l_{max2}$ so that the sample can be assigned to a new label. When classifying the expressions in the testing set, we merge all subclasses into the corresponding main classes and compare them with the ground truths to calculate the overall accuracy.

\section{Experiments}

\subsection{Datasets}
To verify the effectiveness of the proposed model, we conduct extensive experiments on two popular multimodal FER datasets: BU-3DFE \cite{yin20063BU3DFE} and Bosphorus \cite{savran2008bosphorus}.

\noindent\textbf{BU-3DFE} benchmark includes 2,500 3D facial expression scans collected from 100 subjects (56 females and 44 males) for 2D+3D static FER. Each subject has six basic expressions (i.e., anger, disgust, fear, happiness, sadness and surprise) with four intensity levels and one neutral expression. We use the highest two intensity levels for training and testing, as most previous methods did \cite{Li2017Multimodal,Sui2021FFNet}. Some samples of the 2D texture images in BU-3DFE with the highest two levels of intensity are shown in the first two rows of Figure 4.

\noindent\textbf{Bosphorus} contains 4,666 3D scans from 105 subjects ranging from 25 to 35 years old. Only 63 subjects contain all the six basic expressions with one intensity. Examples of 2D texture images with six basic expressions are presented in the bottom row of Figure 4.
\begin{figure}[t]
  \centering
  \centerline{\includegraphics[width=6cm]{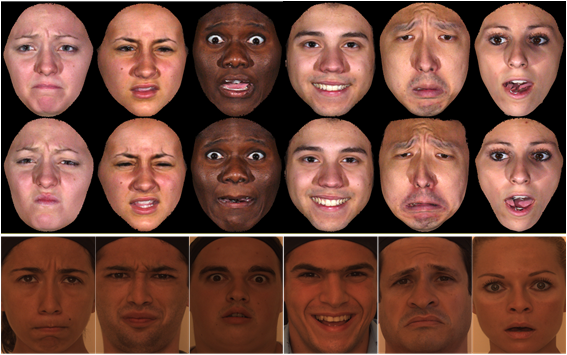}}
  \medskip

\caption{Samples of six basic expressions in each column. Top two rows: 2D texture images with level 3 and level 4 of expression intensity from BU-3DFE; bottom row: 2D texture images from Bosphorus. }
\end{figure}

\begin{table}[t]
\label{1}

\begin{tabular*}{\hsize}{  p{0.13\columnwidth}<{\centering}p{0.18\columnwidth}<{\centering}p{0.26\columnwidth}<{\centering}p{0.26\columnwidth}<{\centering}}

\toprule

 Modality&Fusion&BU-3DFE (\%)&Bosphorus (\%)\\
\midrule

2D    & --   &88.27          & 87.79 \\
3D    & --   &86.32           & 84.17\\
2D+3D & Naive&89.44           & 89.17 \\
\midrule
2D+3D & AF (Ours) & \textbf{90.28} & \textbf{89.72}\\

\bottomrule
\end{tabular*}
\caption{Evaluation of the alternative fusion (AF) strategy. The best results are in bold.}
\end{table}
\begin{table*}[t]
\label{2}

\begin{tabular*}{\hsize}{  p{0.2\columnwidth}<{\centering}p{0.31\columnwidth}<{\centering}p{0.25\columnwidth}<{\centering}p{0.47\columnwidth}<{\centering}p{0.45\columnwidth}<{\centering}}

\toprule

 SF&Fusion Method&Modality&BU-3DFE (\%)&Bosphorus (\%)\\
\midrule

\XSolidBrush & --&2D& 88.27&87.79 \\
\CheckmarkBold & --&2D& 88.89&88.27 \\
\midrule
\XSolidBrush & Naive&2D+3D& 89.44& 89.17\\
\CheckmarkBold & Naive&2D+3D & 90.00& 89.72\\
\midrule
\XSolidBrush & AF&2D+3D & 90.28 & 89.72\\

\CheckmarkBold & AF&2D+3D &\textbf{90.83}&\textbf{90.28}\\

\bottomrule
\end{tabular*}
\caption{ Evaluation of the sample filtering (SF) module. The best results are in bold.}
\end{table*}

\noindent\textbf{Evaluation Protocol} is similar for BU-3DFE and Bosphorus. Specifically, for the BU-3DFE dataset, we randomly select 60 subjects from all the 100 subjects, and only the two highest expression intensities are used for training and testing. The 10-fold cross-validation method is adopted, which means the 60 subjects are randomly and evenly divided into 10 subsets. Each time, we use nine of these subsets for training and the remaining one subset for testing. This is done in turn for 10 times to complete one validation. Such a process is repeated 100 times to obtain stable results, and the average accuracy of these validations is calculated to evaluate the performance. As for the Bosphorus dataset, the only difference from BU-3DFE is that we randomly choose 60 subjects from 63 subjects which include all the six basic expressions with one intensity.
\subsection{Experiment Setting}
The RGB images and depth maps are resized to $224$$\times$$224$$\times$$3$ and $224$$\times$$224$$\times$$1$, respectively. To avoid overfitting, some data augmentation techniques including random erasing, horizontal flipping, and color jittering are employed. We use the DeiT-S in \cite{touvron2020DeiT} pre-trained on ImageNet as the backbone of our MFEViT. At the training stage, we adopt AdamW \cite{loshchilov2018adamw} with betas (0.9, 0.999) to optimize our MFEViT with a batch size of 16. The learning rate is initialized to $4^{-5}$ for both BU-3DFE and Bosphorus in 130 epochs for the cross-entropy loss function. Our sample filtering module is utilized to perform optimization for BU-3DFE and Bosphorus from the 20-th epoch and 35-th epoch, respectively. By default, the threshold of the SF module ($\delta$) and the number of subclasses of each expression ($N$) are set to 0.4 and 5 by default, respectively. All the experiments are conducted on a single NVIDIA RTX 3070 card with PyTorch toolbox \cite{paszke2019pytorch}.
\subsection{Ablation Studies}

\begin{table*}[t]
	\begin{center}
		
		\begin{tabular*}{\hsize}{  p{0.88\columnwidth}<{\centering}p{0.27\columnwidth}<{\centering}p{0.33\columnwidth}<{\centering}p{0.24\columnwidth}<{\centering}}
			\toprule  
			Method & Modality & Feature Extraction & Accuracy (\%) \\
			\midrule
			MS-LNPs \cite{Li20123D}  & 3D & Hand-crafted & 80.14 \\
			DMCMs \cite{Lemaire2013Fully}& 3D & Hand-crafted & 76.61 \\
			iPar-CLR \cite{li2015iPar-CLR} & 2D+3D & Hand-crafted & 86.32 \\
			FERLrTc \cite{fu2019FERLrTc} & 2D+3D & Hand-crafted & 82.89 \\
			\midrule
			DF-CNN \cite{Li2017Multimodal} & 2D+3D & CNN & 86.86 \\
			OT-L$_{p}$L$_{1}$ \cite{wei2018unsupervised} & 2D+3D & CNN & 88.03 \\
			VGG-M-DF \cite{jan2018accurate} & 2D+3D & CNN & 88.54 \\
			FLM-CNN \cite{chen2018fast}& 3D & CNN & 86.67 \\
			DA-CNN \cite{Zhu2019Discriminative} & 2D+3D & CNN & 88.35 \\
			FA-CNN \cite{jiao2019facial} & 2D+3D & CNN &89.11 \\
			GAN-based \cite{zhu2020intensity}& 2D+3D & CNN & 88.75 \\
			OGF$^{2}$Net \cite{Lin2020Orthogonalization} & 2D+3D & CNN & 89.05 \\
            GEMax \cite{jiao20202GEMAX} & 2D+3D & CNN & 89.72 \\
            FFNet-M \cite{Sui2021FFNet} & 2D+3D & CNN & 89.82 \\
			\midrule
			MFEViT (Ours) & 2D+3D & Transformer & \textbf{90.83} \\
			\bottomrule
		\end{tabular*}
        \caption{Comparison results on BU-3DFE. The best accuracy is highlighted in bold.} \label{table2}
	\end{center}
\end{table*}

\begin{table*}[t]
	\begin{center}
		
		\begin{tabular*}{\hsize}{  p{0.88\columnwidth}<{\centering}p{0.27\columnwidth}<{\centering}p{0.33\columnwidth}<{\centering}p{0.24\columnwidth}<{\centering}}
			\toprule  
			Method & Modality & Feature Extraction &Accuracy (\%)  \\
			\midrule
			MS-LNPs \cite{Li20123D} & 3D & Hand-crafted & 75.83 \\
			iPar-CLR \cite{li2015iPar-CLR} & 2D+3D & Hand-crafted & 79.72 \\
			FERLrTc \cite{fu2019FERLrTc} & 2D+3D & Hand-crafted & 75.93 \\
			\midrule
			DF-CNN \cite{Li2017Multimodal} & 2D+3D & CNN &  80.28 \\
			OT-L$_{p}$L$_{1}$ \cite{wei2018unsupervised} & 2D+3D & CNN & 82.50 \\
			Deep Feature Fusion CNN \cite{tian20193d}& 2D+3D & CNN & 79.17 \\
			Multi-view CNN \cite{vo20193d} & 2D+3D & CNN & 82.40 \\
            OGF$^{2}$Net \cite{Lin2020Orthogonalization} & 2D+3D & CNN & 89.28 \\
            GEMax \cite{jiao20202GEMAX} & 2D+3D & CNN & 83.63 \\
            FFNet-M \cite{Sui2021FFNet} & 2D+3D & CNN & 87.65 \\
			\midrule
			MFEViT (Ours) & 2D+3D & Transformer & \textbf{90.28} \\			
			\bottomrule
		\end{tabular*}
        \caption{Comparison results on Bosphorus. The best accuracy is highlighted in bold.} \label{table3}
	\end{center}
\end{table*}
To show the effectiveness of our MFEViT, we conduct ablation studies to evaluate the influence of each component on the overall performance on BU-3DFE and Bosphorus datasets. All the experiments are performed using the same hyperparameters for a fair comparison.

\noindent\textbf{Effectiveness of Alternative Fusion.}
We compare our alternative fusion strategy with the Naive fusion strategy on both BU-3DFE and Bosphorus. The latter strategy refers to copying single-channel depth images into three channels. After that, the three-channel depth map and RGB image are encoded by two independent linear projections and then the average of these two sequences of patch embedding is fed into MFEViT for training or testing. In addition, we also compare the performance of training the network with single-modal data (RGB images or depth maps only).

As reported in Table 1, the training on multimodal data produces a much higher testing accuracy, compared with the training on single-modal data. This is predictable because multimodal data provide more information for the training of the network, thereby improving its generalization ability. Furthermore, our fusion strategy accuracies outperforms the Naive one on both BU-3DFE and Bosphorus datasets. This is because our strategy narrows the gap between multimodal data and RGB images, allowing us to make better use of the pre-trained parameters on large-scale RGB image datasets (e.g., ImageNet).

\noindent\textbf{Effectiveness of Sample Filtering.}
To validate the sample filtering module in our MFEViT, we conduct six experiments under the same setting on both BU-3DFE and Bosphorus. As shown in Table 2, when utilizing our sample filtering module to move noisy samples to subclasses, the testing accuracies are superior to those without distinguishing noisy samples from normal samples, regardless of the fusion methods and data modalities. Specifically, the sample filtering boosts the accuracy of our MFEViT to 90.83\% and 90.28\% on BU-3DFE and Bosphorus respectively, indicating a separate improvement of 1.39\% and 1.11\% compared with those of the model employing the Naive fusion only.

\begin{table*}[t]
	\begin{center}
		
		\begin{tabular*}{\hsize}{  p{0.80\columnwidth}<{\centering}p{0.20\columnwidth}<{\centering}p{0.23\columnwidth}<{\centering}p{0.25\columnwidth}<{\centering}p{0.28\columnwidth}<{\centering}}
			\toprule  
			Method & Modality & Param. No. &Framework& Accuracy (\%)\\
			\midrule
			VGG-M-DF \cite{jan2018accurate} & 2D+3D &  $\approx$327M &multi-branch& 88.54 \\
			DA-CNN \cite{Zhu2019Discriminative} & 2D+3D &  $\approx$463M &multi-branch& 88.35 \\
			OGF$^{2}$Net \cite{Lin2020Orthogonalization} & 2D+3D &  $\approx$137M &multi-branch& 89.05 \\
            FFNet-M \cite{Sui2021FFNet} & 2D+3D &  $\approx$93M &multi-branch& 89.82 \\
			\midrule
			MFEViT (Ours) & 2D+3D &  $\approx$\textbf{23}M &single-branch& \textbf{90.83} \\
			\bottomrule
		\end{tabular*}
\caption{Comparison of parameter number and accuracy on BU-3DFE. The fewest parameters and best accuracy are highlighted in bold.}
\label{table5}
	\end{center}
\end{table*}

\subsection{Comparison with State-of-the-Arts}
\noindent\textbf{Results on BU-3DFE and Bosphorus.}  Table 3 and Table 4 show the performance comparison of our model with other state-of-the-art approaches on BU-3DFE and Bosphorus, respectively. It can be seen that our MFEViT achieves the highest accuracies of 90.83\% and 90.28\%, indicating new state-of-the-art capability. The results on both datasets also demonstrate that the transformer-based features in our model outperform the hand-crafted features and CNN features in other methods, regardless of 3D or 2D+3D FER. \\ \indent Figure 5 gives the confusion matrices on BU-3DFE and Bosphorus, displaying the superiority of our MFEViT. We can see that to a certain extent, the results on BU-3DFE and Bosphorus are similar. Particularly, the accuracies of happiness and surprise are relatively high, while those of fear and sadness are relatively low. This may be because the characteristics of exaggerated happiness and surprise are easier to distinguish, while the attributes of fear and sadness with less muscle deformations are more confusing.
\begin{figure}[t]
\centering
\subfigure[BU-3DFE.]{
\begin{minipage}[t]{0.5\linewidth}
\centering
\includegraphics[width=1.5in]{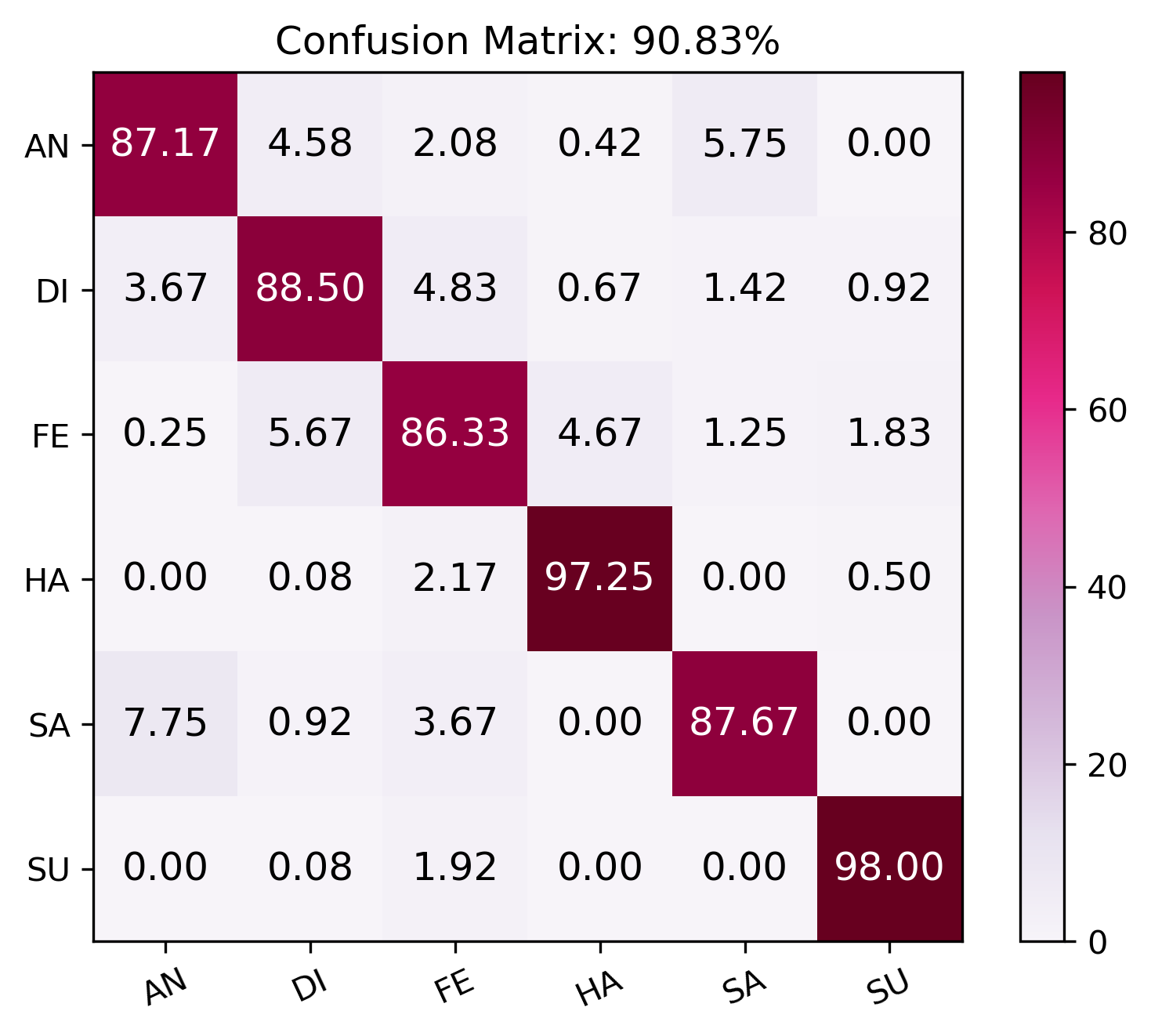}
\end{minipage}%
}%
\centering
\subfigure[Bosphorus.]{
\begin{minipage}[t]{0.5\linewidth}
\centering
\includegraphics[width=1.5in]{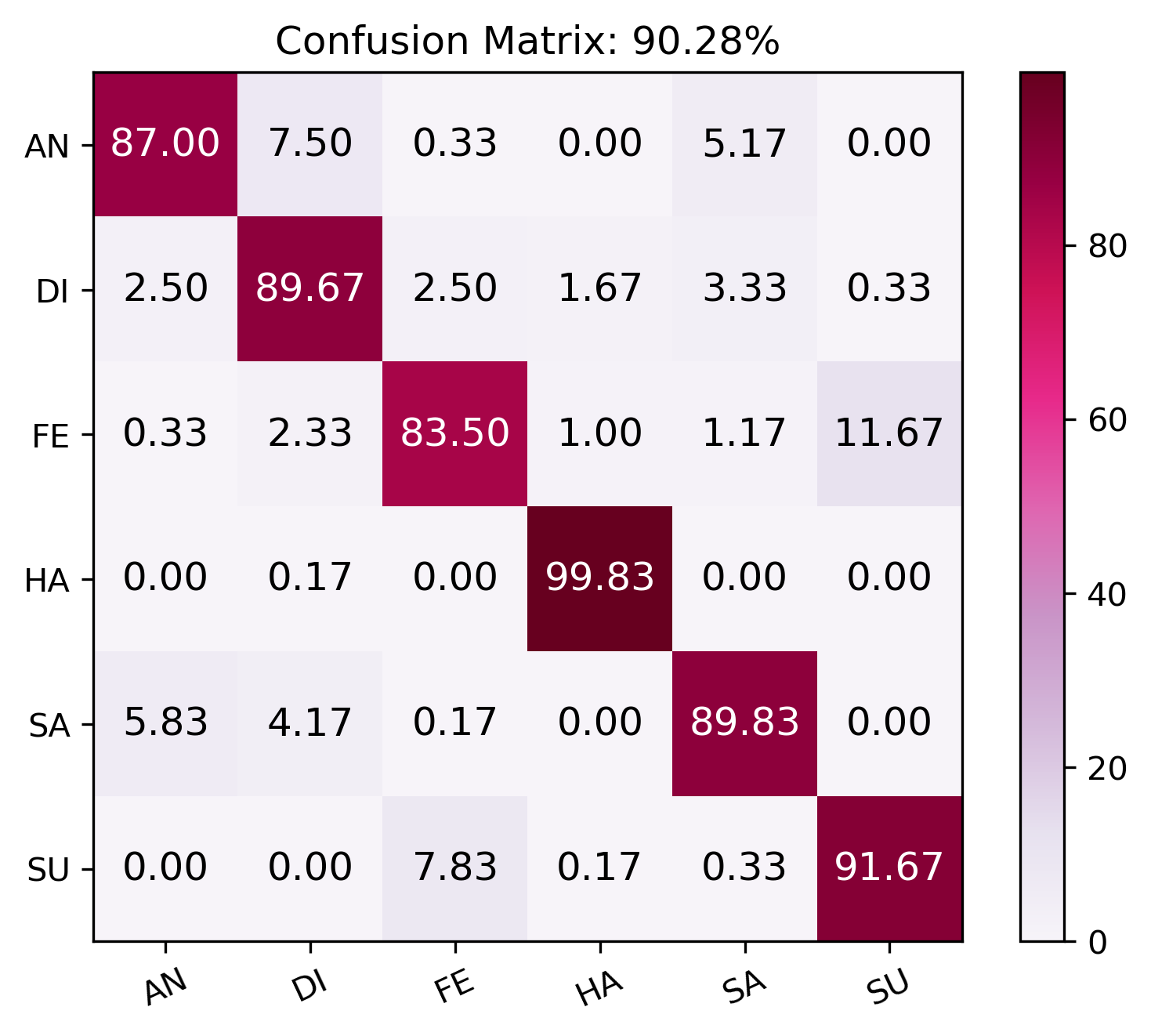}
\end{minipage}%
}%
\centering
\caption{The confusion matrices of our MFEViT on BU-3DFE and Bosphorus. SA, HA, AN, DI, SU, and FE represent sadness, happiness, anger, disgust, surprise, and fear, respectively.}
\end{figure}

\noindent\textbf{Parameter Analysis.} We also carry out comparative studies on BU-3DFE from the parameter perspective. Since the fusion is performed at the data level, our MFEViT consists of only one network branch, which effectively and significantly decreases the amount of parameters. As listed in Table 5, we can see that our MFEViT reaches the best accuracy (90.83\%) with the smallest amount of parameters, compared with other multi-branch models.

\begin{figure}[!t]
  \centering
  \centerline{\includegraphics[width=9cm]{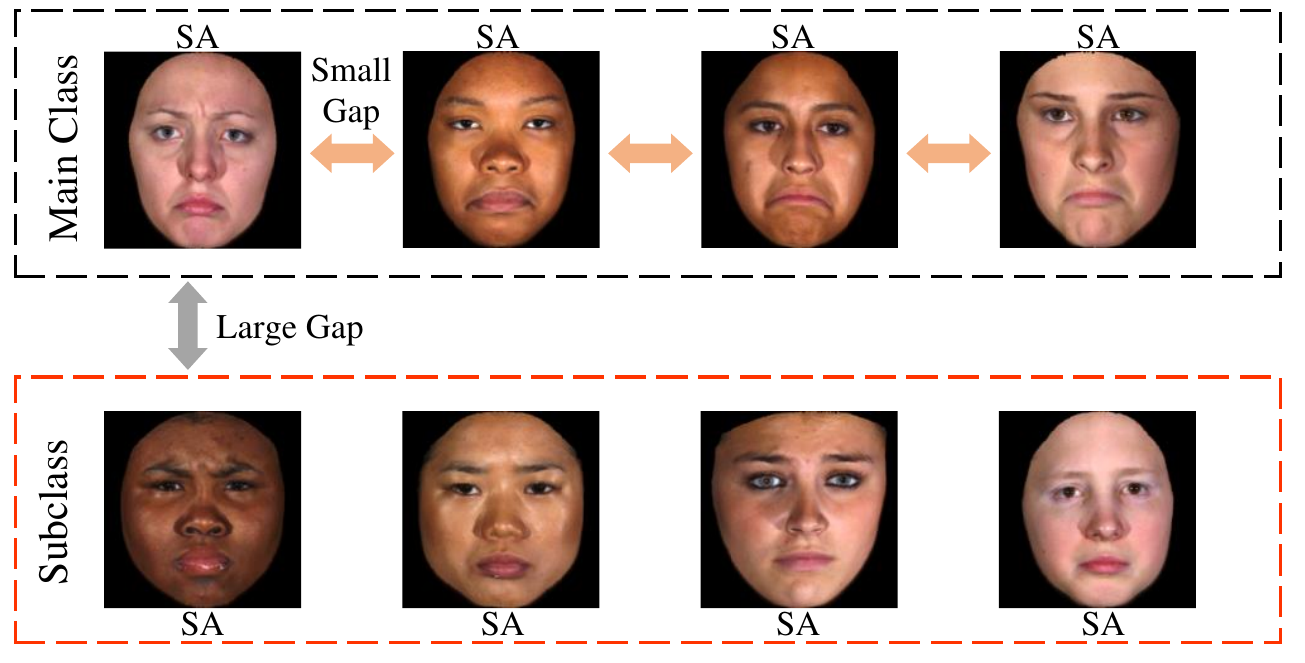}}
  \medskip

\caption{Visualization of the main class and subclasses of sadness figure in the BU-3DFE dataset. }
\end{figure}
\subsection{Visualization}
Our sample filtering module relabels noisy samples into subclasses of the original expression, which reduces the interference of noisy samples to the main expression class during the training process. To visually demonstrate this superior capacity, we visualize some figures belong to the class of sadness from BU-3DFE in Figure 6. The examples in the bottom box are noisy samples relabeled to the subclasses by our sample filtering module, and those in the top box are the normal samples in the main class. It is obvious that the distance between the normal samples is significantly smaller than the distance between the normal samples and the noisy ones. The result also shows that our module plays an important role in filtering the noisy samples, thus better solving the problem of large intra-class distance in FER.

\section{Conclusion}
In this work, we develop a robust lightweight pure transformer-based network named multimodal facial expression vision transformer for multimodal (2D+3D) FER. To the best of our knowledge, this is the first pure transformer-based framework in the multimodal FER field. Specifically, we introduce an alternative fusion strategy to fuse RGB images and depth maps, which narrows the gap between RGB images and multimodal data, and also greatly reduces the amount of parameters. At the training stage, we design a novel sample filtering module to move the noisy samples to the subclasses of each expression, which effectively decreases the impact of noisy samples on network generalization ability. Extensive experiments demonstrate that our MFEViT outperforms other state-of-the-art methods on BU-3DFE and Bosphorus in terms of both accuracy and amount of parameters.

\bibliographystyle{IEEEbib}
\bibliography{refs}

\end{document}